# Critical Insights into Leading Conversational AI Models

Urja Kohli[1,a], Aditi Singh[2,b,] Arun Sharma[3,c]

Corresponding Author: Urja Kohli

[1] Department of Mechanical and Automation Engineering, Indira Gandhi Delhi Technical University for Women, Delhi, India

[2] Department of Electronics and Communications Engineering, Indira Gandhi Delhi Technical University for Women, Delhi, India

[3] Department of Information and Technology, Indira Gandhi Delhi Technical University for Women, Delhi, India

[a] urja090btmae22@igdtuw.ac.in, [b] aditi003btece23@igdtuw.ac.in, [c] arunsharma@igdtuw.ac.in,

**Abstract**

Big Language Models (LLMs) are changing the way businesses use software, the way people live their lives and the way industries work. Companies like Google, High-Flyer, Anthropic, OpenAI and Meta are making better LLMs. So, it's crucial to look at how each model is different in terms of performance, moral behaviour and usability, as these differences are based on the different ideas that built them. This study compares five top LLMs: Google's Gemini, High-Flyer's DeepSeek, Anthropic's Claude, OpenAI's GPT models and Meta's LLaMA. It performs this by analysing three important factors: Performance and Accuracy, Ethics and Bias Mitigation and Usability and Integration. It was found that Claude has good moral reasoning, Gemini is better at multimodal capabilities and has strong ethical frameworks. DeepSeek is great at reasoning based on facts, LLaMA is good for open applications and ChatGPT delivers balanced performance with a focus on usage. It was concluded that these models are different in terms of how well they work, how easy they are to use and how they treat people ethically, making it a point that each model should be utilised by the user in a way that makes the most of its strengths.

**Highlights**

- Proper study of five prominent LLMs: Gemini, DeepSeek, Claude, GPT and LLaMA.
- Comparison of each AI model based on accuracy, ethics and ease of use.
- Choice of the right LLM depends on the individual use case to optimise the design and features of each model effectively.

## 1.1 Introduction

Artificial Intelligence (AI) has advanced rapidly over the last few decades. While early AI focused on rule-based systems and symbolic reasoning, it has since moved to networks and deep learning, which enable models to handle intricate data patterns. Natural Language Processing (NLP) is a prominent use of AI in language generation and comprehension. It dates back to the early days of NLP, when statistical models were used. In 2017, the transformers made the transition to deep learning possible. Transformers have directly contributed to a significant advancement in language modeling and, in turn, the emergence of LLMs.

Large models with billions of parameters that have been trained on massive text corpora are unquestionably LLMs. Text generation, reasoning, summarization, coding and translation are among the primary LLM skills. A growing number of chatbots, assistants and search business tools are available on the market as a result of the paradigm shift from narrow natural language processing to general-purpose conversational AI. LLMs are currently influencing how users interact with enterprise software, industries and daily life. Leading companies are currently in a perfect race to develop better LLMs, including Google's Gemini, High-Flyer's DeepSeek, Anthropic's Claude, OpenAI's GPT models and Meta's LLaMA. It is important to note that every model is based on a unique architecture, set of training data and set of performance objectives.

As a result, it becomes necessary to examine the strengths and limitations of each model in terms of accuracy, bias, usability and ethics and to compare them in order to provide researchers, developers and companies with a better understanding. This brings us to the main goal of this research paper, which is to compare and contrast five important LLMs: Google's Gemini, High-Flyer's DeepSeek, Anthropic's Claude, OpenAI's GPT models and Meta's LLaMA.

## 1.2 Review Methodology

This study selected the five most popular leading LLMs[19], [4] and [16]. Each model underwent 6-7 evaluations to maintain the unbiased nature of the research. The methods used include systematic literature survey along with the designed case studies. This included extracts from the published journal studies and conferences papers. Grey literature such as articles and blogs have also been considered. Various comparison variables were chosen to accomplish the goals of this study. The comparison variables consisted of language comprehension, content development[17], performance considerations[5], scalability and architecture & planning. Each model was analyzed based on its unique prompts which advocated for its various strengths and possible use cases in the multiple industries. Official manuscripts and publications were considered to develop the appropriate questions. The data was systematically analyzed to evaluate the performance of the models under specific circumstances. Performance & Accuracy, Ethics & Bias Mitigation and Usability & Integration were the three main criteria to compare the models.

## 2. Literature Review

### 2.1 Introduction to Conversational AI Models

In a study, Zang et al. conducted a comprehensive evaluation of the capabilities of ChatGPT in scientific research. It analyzed 50 abstracts and 4 research articles generated by the model and reviewed by 23 experts[1]. The researchers discovered that ChatGPT lacked the strength to deal with the technical parts of the study. However, it could be used to generate good academic material if provided with well compiled suggestions. The reviews' ratings provide us a better idea of how AI can and can't be used in academic writing in the actual world. The five biggest drawbacks with using ChatGPT were hallucinations, originality issues, toxicity risks, privacy concerns and sustainability[2]. Hallucination occurs when the model gives information that sounds correct but is wrong. This happens because the training data is varied, the model's algorithmic limits and the way prompts are set up.

### 1.2.3 A Comparison of the Best Conversational AI Models

Current research seeks to comprehend the distinct advantages and disadvantages of many conversational AI models via comparative analysis. A comparison research shows that Google's Gemini does a better job of sentiment analysis, getting facts right and giving answers that are relevant to the situation than ChatGPT. Conversely, ChatGPT is superior at broader conversational tasks [3]. Gemini is used in construction, education, healthcare and finance because it can debug code, optimise it and analyse data.

A research looked at the answers to 1002 questions and rated four top chatbots in 27 different areas[5]. The findings showed that GPT-4 was the most accurate, with an accuracy of 84.1%, while Claude was the least accurate, with an accuracy of 64.5%. GPT-4 did very well on language comprehension and reasoning, getting 92.24%. Claude, on the other hand, stood out for his capacity to moderate bias. A study that compared LLaMA to GPT-3 and other bigger models showed how well it worked [4]. It is trained on 1.4 trillion tokens from GitHub, Wikipedia, ArXiv and CommonCrawl. LLaMA-13B performed better than GPT-3 despite having a smaller size compared to other models. LLaMA-65B managed to match the level of the larger models like PaLM-540B and Chinchilla-70B. However, research has found out that the model showed bias against women, old generation and religious communities. A full study was conducted of its energy usage by interpreting a PUE of 1.1 and a carbon intensity factor of 0.385 kg $CO_2$ per kWh in the US. These results suggest the significance of LLaMA in open-source LLMs. An experiment assessed the replies to 1002 questions to rate four of the best chatbots in 27 different areas [5]. The results show that Claude was 64.5% accurate and GPT-4, which performed the best, was 84.1% accurate. GPT-4, with a score of 92.24%, exhibited proficiency in language comprehension and reasoning. There was a strong link (0.82) between ChatGPT and GPT-4, which could mean that similar methods have been employed to solve problems[6].

### 1.2.4 Core Capabilities

The capabilities of various LLMs have been reviewed on the basis of information extraction, creativity and performance. The strength and limitations of ChatGPT were revealed through an analysis on fourteen fundamental extraction tasks. The results suggested that its performance was significantly affected by complex targets and irrelevant context[7]. The relationship between the subject and the object was difficult for it to understand. This raised concerns over its reliability in activities that involved data annotation.

The effectiveness of DeepSeek's design, which included Mixture-of-Experts and Multi-Head Latent Attention, was assessed [25]. It performed exceptionally well on benchmarks such as miniF2F. It also excels in code production, structured reasoning and medical diagnostics. DeepSeek offers better scalability and efficiency at a cheaper training cost than proprietary models like ChatGPT and Gemini. However, it falls short in areas like financial forecasts, creative work and linguistic subtleties, particularly in different forms of English writing.

The open-source nature of DeepSeek allows collaboration of AI. Although, it also raises issues regarding safety and ethical considerations. DeepSeek represents a promising and cost-effective option within the LLM domain with room for improvement in creativity and bias management.

A research study also examined the reasoning capabilities of various large language models. The results showed that they excelled at reasoning tasks with an average score of 66%. The models excelled in commonsense reasoning (82.64%) and temporal reasoning (82.70%) whereas they struggled with spatial reasoning (33.75%) and physical reasoning (51.92%)[20]. The conclusion emphasized the gap between AI's reasoning capabilities and human reasoning.

Additionally, a research experiment divided questions into three categories based on their difficulty and how well models handled them. Easy questions (441 in total) were a breeze for all the models. When it came to medium-difficult level questions (416), only one or two models answered them correctly[10]. But the challenge came with the tough ones (129), where even GPT-4 struggled. This classification method as summarised in Figure 1 provides a systematic way to evaluate and compare models as they evolve.

| Feature | Gemini [3],[6] | ChatGPT [3],[6] | LLaMA [4] | DeepSeek [22] |
|---|---|---|---|---|
| Sentiment Analysis | ✓ | ~ | ✗ | ✗ |
| Factual Accuracy | ✓ | ~ | ✓ | ✓ |
| Context Response | ✓ | ✓ | ✗ | ~ |
| Code Debugging | ✓ | ✗ | ✗ | ✓ |
| Data Analysis | ✓ | ✗ | ✗ | ✓ |
| Image Recognition | ✓ | ✗ | ✗ | ✗ |
| Reasoning | ~ | ✓ | ~ | ✓ |
| Comprehension | ~ | ✓ | ~ | ✓ |
| Ethics | ✓ | ✓ | ~ | ~ |

*Figure 1 Comparative Analysis of Gemini, ChatGPT, LLaMA and DeepSeek [3],[4],[6],[22]*

**1.2.5 Specialized Applications of Conversational AI**

Conversational AI has increasingly become a transformative tool across industries, particularly in education and healthcare. A research study emphasizing ChatGPT's Role in Ophthalmology[11] provided an in-depth look at ChatGPT's understanding of ophthalmology that was included the examination of its performance in different subspecialties. Results indicated that the legacy model performed well in general medicine (75%), core fundamentals (60%) and cornea-related topics (60%). However, it struggled in specialized areas like neuro-ophthalmology (25%), glaucoma (37.5%) and pediatrics and strabismus (42.5%)[11]. The model's performance improved with subsequent updates. This suggests that further refinements can enhance its utility in medical fields. These findings underscore both the promise and limitations of large language models in advancing specialized medical knowledge and their evolving potential for clinical support.

*Table 1 Summary of ChatGPT's Specialised Application [11],[12]*

| Feature | ChatGPT |
| --- | --- |
| Ophthalmology | Strong in general medicine, weak in specialties [11] |
| Medical Use | Needs refinement for clinical support [11] |
| Academic Integrity | Concerns with AI-generated content detection, only 70% accuracy with AI detectors [12] |

Beyond medical applications, the rise of AI has raised new challenges. According to the findings of a study focused on the concerns about the detection of AI-generated content in academic settings[12], the current plagiarism detection tools still have major drawbacks. AI detectors like RoBERTs and even human judgment could identify AI-generated abstracts only 70% of the time. GPTZero is based on an innovative approach that looks at randomness and "burstiness," and it is found that it struggles to make accurate classifications[12]. These instances underline the growing challenge of safeguarding academic integrity in an era where AI-driven content generation is becoming increasingly pervasive.

There are concerns about AI's impact on academic integrity but it is also getting popular as a potential educational tool. The analysis of an emerging technology report on Gemini's ability as an educational technology in creating customized and multimodal learning experiences highlights that Gemini 1.0 Ultra is highly capable of analyzing code, solving logic-based tasks and tackling challenging math problems. Gemini 1.5 Pro can parse up to 100,000 lines of code and demonstrates a sophisticated understanding of complex scientific concepts [13]. Gemini can create content that is appropriate for individual learning and can adjust to various communication methods. Therefore, it is a reliable instrument for individualized instruction. Researchers emphasize on the need for clear ethical guidelines to ensure that the use of AI in education remains effective and equitable.

*Table 2 Summary of Gemini's Specialised Application [13],[14]*

| Feature | Gemini |
|---|---|
| Educational Potential | Strong in personalized learning, can analyze code, solve logic and math problems [13] |
| Performance in Code and Logic Tasks | Strong in code analysis and problem-solving [13] |
| Adaptability to Communication Styles | Adapts well to learning styles [13] |
| Ethics | Needs ethical guidelines [13] |
| Information Sector Impact | Improves voice recognition and info access [14] |
| Customer Service | Enhances user satisfaction [14] |

Numerous industries are changing as a result of Gemini's growing influence. Increased competition in the technical industries led to advancements in natural language processing and voice recognition, according to the systematic assessment of the Gemini's impact on the information sector[14], a report that assesses Google's development of Gemini as a competitive force within the information business. However, it further states that Gemini distinguishes itself from other AI assistants like Apple's Siri and Microsoft's Xiaoice[14]. Better information access, faster material processing and personalized recommendations are some of the characteristics that make Gemini unique. Additionally, through accurate and easily accessible information, the research demonstrates how this competition is raising the bar for customer service standards that encourage better user interaction and raise overall satisfaction.

### 1.2.6 Ethical Considerations and Challenges

The review of Ethics and the Road Ahead for LLMs [15] points out the most important issues that need to be looked at further in the development of large language models (LLMs). The first is the possible use of ensemble methods, which integrate the best parts of several models. Second, it stressed how important it is to improve your thinking skills. Third, it stressed the need to maintain high levels of technical performance while still following moral rules. Fourth, it emphasised the need for robust evaluation systems and the imperative for moral issues to be regarded with similar importance as scientific progress [15].

A research found that DeepSeek-R1 is a reasoning-optimized LLM that does very well in math, coding and logic tasks after going through multi-stage reinforcement learning. It also shows how cultural and linguistic prejudices might be limited since it only looks at English and Chinese. The paper talks about moral issues including justice, prejudice and the fact that there isn't enough human supervision in the reinforcement learning process. DeepSeek-R1 uses RL, fine-tuning, benchmarking and knowledge distillation as its validation methods. These have not been particularly helpful in solving the problems it has with safety and explainability. To reduce these dangers, the people that created DeepSeek-R1 have suggested ways to make the model's multilingual data bigger, do bias audits, add human-in-loop mechanisms and use third-party validation to make sure that the model is morally responsible.

In conclusion, DeepSeek-R1's clear documentation and ethical standards make it a good choice for responsible AI research. But it is important to keep working on its ethical flaws if technology is going to be used in a way that lasts [24].

### 1.2.7 Advancing the Field: Integration and Future Prospects

To advance the field of large language models, it is essential to examine model integration and future prospects. Research on Integrating Models has demonstrated that using many language models can significantly increase accuracy, with one model providing the correct response. But when every model agreed, accuracy dropped to 44% and when none did, it dropped to 7%. The combined model method demonstrated the potential of ensemble approaches to improve performance with an astonishing 97.6% accuracy on language-related tasks.

Furthermore, the Multi-Model Integration Framework demonstrates how combining models can benefit from their unique advantages, overcoming individual limitations and improving performance, particularly in complex problem-solving and specialized professions [17]. In tasks that require sophisticated thinking, this multi-model approach improves accuracy and efficiency.

Key barriers to increase reasoning, guarantee factual correctness, improve model evaluation and resolving ethical difficulties are identified by Chitimoju, S. in his study, The Evolution of Large Language Models: Trends, Challenges and Future Directions. The study promotes continuous improvements in integration methods, stressing the necessity of a well-rounded strategy that combines technological innovation with responsible application [18].

### 1.2.8 Cutting-Edge Innovations

Multimodal AI has advanced significantly with the exploration of novel models, beginning with Google's new Gemini model family (Ultra, Pro and Nano variants). Gemini Ultra is a flexible tool with potential uses in education since it performs exceptionally well on tasks involving speech, graphics, video and text. Complex competitive programming issues are easily solved by the AlphaCode 2 version of Gemini [19]. Longer context processing, less hallucinations and trustworthy knowledge retrieval are further noteworthy characteristics. According to research, Gemini can successfully handle even "unanswerable" enquiries because of its thorough training on a variety of data sources. LLaMA 3.1 provides a comprehensive performance comparison with industry leaders such as GPT-4, having been assessed across 150 benchmarks, even though Gemini exhibits amazing capabilities. LLaMA 3.1 was trained on 16,000 H100 GPUs and quantised to reduce data processing from 16-bit to 8-bit formats for increased efficiency [20]. Its aptitude for reasoning, multilingualism and coding challenges are among its strong points. However, challenges persist in areas such as preserving data quality, resource optimization and addressing the "black-box" nature of decision-making, pointing to areas for further development.

## 2. Discussions
### 2.1 Review-based Comparative Analysis

A thorough side-by-side comparison of the top five Large Language Models (LLMs) is shown in the following tables. Key components like model architecture, training data and contextual understanding are highlighted in table 3a which how these fundamental decisions affect each model's scalability, multimodal capabilities and capacity to carry on coherent discussions. Understanding these elements is imperative to deciding which model is most appropriate for a certain study or practical use.

Table 3a: Architecture, Training and Contextual Abilities

| Feature | ChatGPT | Claude | Gemini | LLaMA | DeepSeek |
|---|---|---|---|---|---|
| Model Architecture | Transformer-based, scalable, RLHF fine-tuned for alignment | Transformer-based, safety-focused with dynamic ethical thresholds | Transformer-based, excels in multimodal tasks and Google integration | Transformer-based, open-source, efficient | Transformer-based, optimized for research, supports base and chat variants |
| Training Data | Books, articles, web pages | Public web pages and human feedback | Multimodal data: web docs, books, code, Google data | 1.4 Trillion tokens, public web data (CommonCrawl) | 2 Trillion tokens: web data, code, curated datasets |
| Contextual Understanding | Maintains context in extended chats | Strong safety in conversational context | Strong contextual understanding | Lacks coherence in long chats[10] | Maintains reasonable context; optimized for tasks and code generation |

As shown in Table 3b, there is an evident difference between the models in terms of safety and ethics. ChatGPT uses Reinforcement Learning from Human Feedback (RLHF) values whereas Gemini uses secured content filtering to properly align outputs with human. It was also found that LLaMA weighs on its open source availability and flexibility while DeepSeek offers minimal in-built safety features, despite being lightweight and accessible. These factors significantly affect the risks and realistic deployment strategies of each model.

Table 3b: Linguistic, Ethical, Efficiency and Tool Features

| Feature | ChatGPT | Claude | Gemini | LLaMA | DeepSeek |
|---|---|---|---|---|---|
| Linguistic Diversity | Many languages, strong English | Multilingual, optimized for English | Strong multilingual support [7] | Broad language support via fine-tuning | Strong English, emerging multilingual via fine-tuning |
| Ethical Consideration | RLHF safeguards, OpenAI filters | Ethical AI, Anthropic principles | Google safety filters | Open-source, community-driven safety | Limited safety compared to closed models |
| Latency and Efficiency | Fast, scalable | Moderate latency, safe responses | Efficient, tuned for Google ecosystem | High efficiency, adaptable | Lightweight (e.g., DeepSeek-Coder), low inference cost |
| Tool Integration[23] | Integrated: DALL-E, browsing, Python | Limited tool integration | Google product integration | Open-source, framework compatible [11] | Easy integration with development frameworks |

Lastly, the primary advantages and disadvantages of each model have been summarised in the Table 3c so the users can maximize the potential of each model while being aware of its limitations.

Table 3c: Strengths and Weaknesses

| Feature | ChatGPT | Claude | Gemini | LLaMA | DeepSeek |
|---|---|---|---|---|---|
| Key strengths | Versatile, strong conversational abilities, broad use cases | High ethical standards, safe output | Strong Google integration, innovative features | Open-source, research adaptable | Strong for coding and research, open access and reproducible |
| Known Weaknesses | lacks specificity and accuracy | Less creative than other models | Limited performance data in varied tasks | Lacks coherence in long dialogues | Limited safety filters, lacks multimodal support |

Building upon these qualitative insights, the following section introduces a quantitative analysis of six core capabilities [3] including Accuracy, Reasoning Ability, Understanding Context, Creativity, Handling Ambiguity and Domain Knowledge, as illustrated in Figure 2. This assessment provides a more detailed insight of how well each model performs on various cognitive tasks. A scale of 0 to 2 was used to rate these models. Both ChatGPT and Gemini consistently score higher than 1.5 and do well in the majority of categories. ChatGPT performs exceptionally well in Creativity, achieving nearly a perfect score of 2.0, while Gemini stands out in Accuracy, scoring nearly 1.9.

Although it performs well, LLaMA typically scores a little lower than its competitors, especially in Reasoning Ability, but it still manages to produce respectable results in other domains. With the highest score in the category of Handling Ambiguity, DeepSeek indicates an important strength in that area. Gemini's accuracy, ChatGPT's creativeness, DeepSeek's ability to manage ambiguity, Claude's stability and LLaMA's strong overall performance across features are all revealed in this comparison.

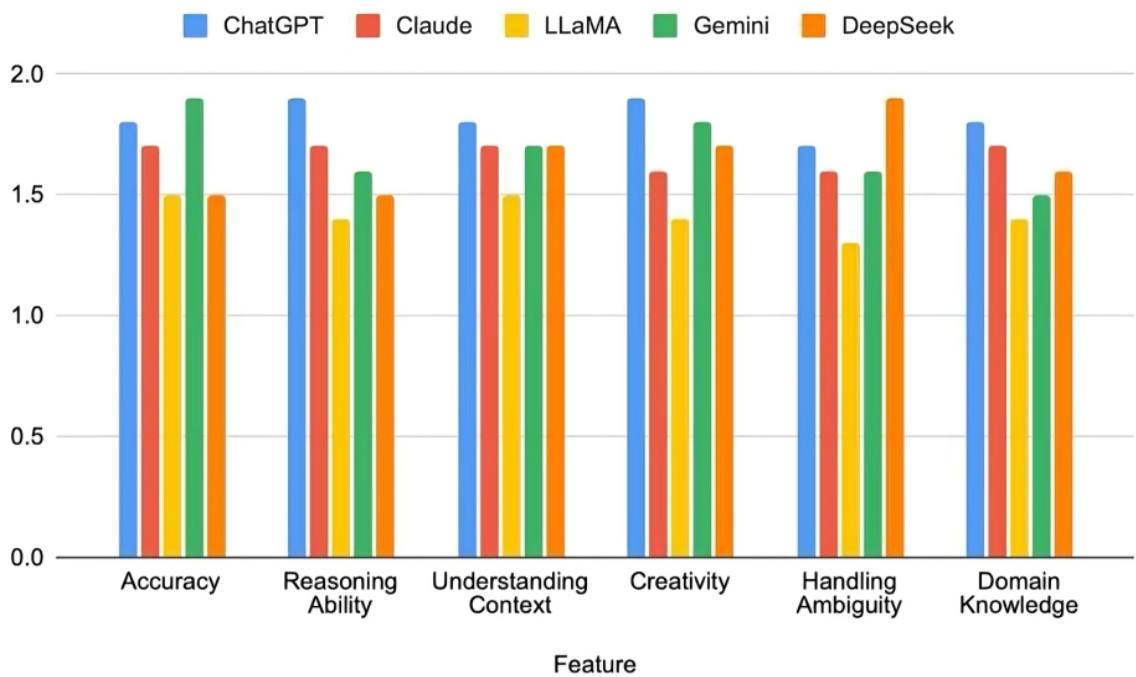

*Figure 2 Cognitive Capabilities Comparison Across Large Language Models (Scale 0-2) [22][30] [31] [32]*

While Figure 2 illustrates each model's performance across fundamental cognitive functions, it is as crucial to examine how well they function in real-world deployment scenarios. The emphasis in Figure 3 is shifted to performance-related characteristics including speed, task consistency and scalability, which have a direct bearing on user experience and practicality. The fundamental features of these five prominent language models [15], namely ChatGPT, Claude, LLaMA, Gemini and DeepSeek, are contrasted below along six important dimensions. With an average score of about 1.8, ChatGPT continuously performs well, particularly in Consistency and Speed of Response [28]. Gemini performs well in User Intent Understanding, scoring close to 1.9, while Claude performs equitably at 1.6, with Task Performance and Consistency being his strongest areas. Although it performs consistently in every category, LLaMA has somewhat lower scores in Scalability and Error Correction [10]. With some of the highest scores in Task Performance and Scalability, DeepSeek exhibits strengths in these domains.

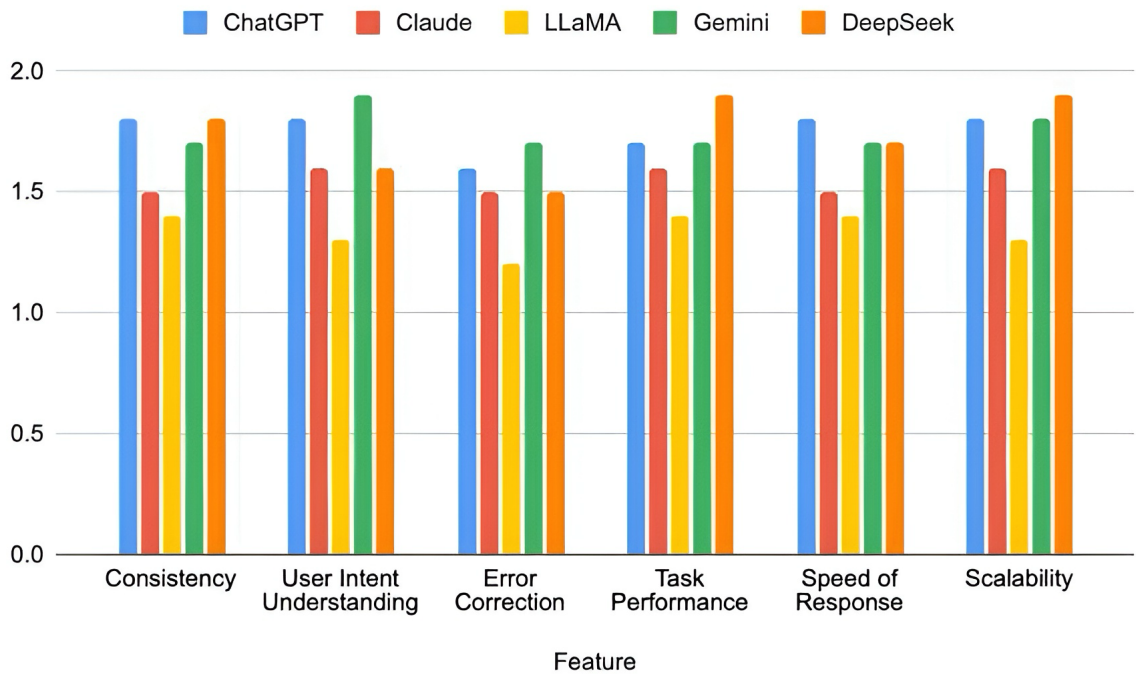

*Figure 3 Assessment of Core Capabilities in Modern LLMs (Scale 0-2) [22][30] [31] [32]*

Conclusively, the graphical representation in Figure 3 indicates the varying strengths of each model, where performance gaps reflect their distinct training approaches and design philosophies. Expanding upon this comparison, the following section builds upon a case study that was carried out for a deep evaluation of these models in practical scenarios.

**2.2 Case Study: Prompt-Based Comparative Evaluation of Leading Conversational AI Models**

This case study presents a prompt-based evaluation of the five leading AI models including DeepSeek, ChatGPT, Gemini, Claude and LLaMA. Three unique prompts were developed based on different assessment criteria. Firstly, the models were asked to describe the primary causes of climate change and propose a clear solution based on evidences. Next, each model was prompted to give a detailed strategy for preventing prejudice in hiring decisions using AI. Lastly, the models had to arrange a collection of sales data as a markdown table and give a justification for their formatting decisions. Each prompt was submitted individually to all five models and the responses were compared in tabular form. In order to prevent plagiarism, key parts from each output were paraphrased while making sure that the core content was intact.

The results and observations from each prompt-based evaluation, beginning with the first prompt, are listed below.

**Prompt 1:** Explain the main causes of climate change and suggest one solution, with evidence. Keep your answer clear and focused.

This specific prompt was selected to assess the models on the performance and accuracy parameter because it calls for the capacity to prioritize pertinent details, retain accurate factual information and synthesize it logically. Additionally, it provides information on the differences between the models in terms of reasoning clarity, factual accuracy and depth of knowledge.

Table 4: Comparative Analysis of Model Responses to Performance and Accuracy Prompt

| Model | Key Response | Interpretation |
|---|---|---|
| DeepSeek | Identifies greenhouse gases ($CO_2$, $CH_4$) from fossil fuels, deforestation and agriculture as main causes. Suggests switching to renewables, citing a study showing up to 70% emission reduction by 2050. Notes renewables are now often cheaper than fossil fuels. | • Detailed, evidence-based answer with quantitative data and real-world examples.<br>• Highlights economic feasibility.<br>• No direct source links in output. |
| ChatGPT | Attributes climate change to increased greenhouse gases from human activities (fossil fuels, deforestation, industry). Recommends transition to renewables, referencing IRENA's estimate of 70% emission reduction by 2050. | • Clear, concise and well-structured.<br>• Cites a reputable organization (IRENA) for evidence.<br>• No direct source links in output. |
| Gemini | Explains enhanced greenhouse effect from fossil fuels, deforestation, agriculture and industry. Advocates rapid shift to renewables, referencing international consensus and IRENA's 90% renewables by 2050 target. Provides cost and deployment evidence. | • Comprehensive, structured and includes international targets.<br>• Provided source links at the end of the response. |
| Claude | Lists fossil fuel combustion, deforestation, agriculture and industry as main causes. Recommends scaling renewables, citing IEA data on cost drops and real-world examples (Denmark, Costa Rica). Notes job creation and energy security benefits. | • Strong on quantitative evidence and real-world feasibility.<br>• Provided source links at the end of the response. |
| LLaMA | Cites fossil fuels, deforestation, industry and agriculture as main causes. Suggests transition to renewables, referencing the Paris Agreement and EU net-zero targets. Notes that renewables are key to limiting warming to 1.5°C | • Concise, policy-aware and references international agreements.<br>• Provided two source links at the end of the response. |

The comparison of the models' answers to this prompt has been summarized in Table 4. In this case, the models' capacity to accurately and pertinently summarise important study findings was evaluated. Overall, the models performed well; some concentrated on clarity and conciseness, while others included more specific proof or real-world instances. Notably, Claude delivered thorough information with significant ethical implications, whereas Gemini gave organised, fact-based answers.

**Prompt 2:**
*How can a company using AI to screen job applicants ensure a fair, unbiased process? List steps and justify your recommendations.*

This particular prompt was chosen to assess the ethics and bias mitigation models because it calls for a thorough comprehension of AI systems' fairness as well as the capacity to pinpoint actions that may be taken to lessen bias. The model's ethical sensitivity and the extent to which it supports the actions suggested to guarantee a fair process are the main points of emphasis.

Table 5: Comparative Analysis of Model Responses to Ethics and Bias Mitigation Prompt

| Model | Key Response (Paraphrased) | Interpretation |
|---|---|---|
| DeepSeek | Recommends auditing training data for bias, using explainable AI tools, regularly testing for disparate impact, ensuring human oversight with diverse review panels and seeking third-party bias certification. Each step is justified with research evidence (e.g., Harvard Business Review, MIT, Stanford, LinkedIn, Forrester). | • Highly evidence-based, detailed and practical.<br>• Cites multiple studies and industry reports for each recommendation,<br>• Strong focus on transparency and real-world validation. |
| ChatGPT | Advises using diverse, representative training data, conducting regular bias audits, ensuring transparency and explainability, including human oversight, establishing clear selection criteria and providing AI ethics training to HR staff. Each step is justified with a focus on fairness, accountability and best practices. | • Comprehensive and structured<br>• Includes both technical and human factors.<br>• Focuses on transparency, regular audits and ethical training<br>• Does not cite specific studies or provide source links. |
| Gemini | Suggests diversifying and auditing training data, focusing on skills and competencies (not demographics), implementing explainable AI, incorporating human oversight, conducting regular bias audits, providing candidate opt-out and feedback mechanisms and training HR professionals on AI bias. Each step is justified with practical and ethical reasoning. | • Thorough and policy-aware<br>• Strong emphasis on transparency, candidate rights and ongoing monitoring.<br>• Includes justifications for each step and references to international standards<br>• Cites source links in the response. |
| Claude | Recommends auditing training data, using diverse datasets, implementing regular bias testing, removing or anonymizing biasing information, establishing human oversight, monitoring outcomes and providing transparency and appeals processes. Justifies each step by explaining how it addresses bias and promotes fairness. | • Balanced approach that combines both technical safety and human oversight.<br>• Notes the importance of transparency and appeals<br>• Provides source links for tools and best practices. |
| LLaMA | Outlines steps including data audit and bias detection, algorithmic transparency, fairness metrics, human oversight, continuous monitoring, diverse development teams, candidate feedback and compliance with regulations. Each step is justified as essential for minimizing bias and ensuring fairness. | • Concise and systematic<br>• Covers both technical and organizational measures.<br>• Considers compliance and feedback<br>• Provides a broad framework for ongoing bias mitigation.<br>• No explicit source links mentioned. |

The comparison of the models' answers to this prompt has been summarized in Table 5. In this case, the capability of each model to handle moral dilemmas in a situation where a recruiting manager had to make an objective choice was assessed. The results show the various model approaches differed; some prioritized data audits, while others included human oversight and third-party certifications. DeepSeek provided the most evidence-based response, citing specific research, while Claude and Gemini focused on practical steps for equity and transparency.

**Prompt 3:**
*Format this sales data as a markdown table and explain your formatting choices: Q1: $12,000; Q2: $15,500; Q3: $14,200; Q4: $16,800. Keep your explanation brief and clear.*
This particular prompt was chosen to test the models' accuracy and clarity in data formatting as well as their ability to give understandable, approachable explanations. Table 6 provides a summary of the models' answers to this prompt.

Table 6: Comparative Analysis of Model Responses to Usability and Integration Prompt

| Model | Key Response (Paraphrased) | Interpretation |
|---|---|---|
| Deepseek | Table with headers "Quarter/Revenue", left-aligned text, right-aligned numbers, consistent use of $ and commas, minimalist design. | • Professional<br>• Scannable presentation<br>• Explicit formatting choices |
| ChatGPT | Simple two-column table ("Quarter", "Sales $"), clear headers, separator row, neat alignment. | • Clarity<br>• Structure<br>• Neatness in markdown |
| Gemini | Table with "Quarter/Sales Revenue", uses pipes and hyphens for markdown, explains alignment with colons and uniquely offers an "Export to Sheets" option. | • Shows markdown syntax details<br>• Includes technical explanation<br>• Adds a practical export feature |
| Claude | Table with left-aligned headers, right-aligned revenue, consistent currency, concise quarters, notes Q3 dip. | • Focus on readability<br>• Professional formatting<br>• Includes data trend insight. |
| LLaMA | Markdown table with "Quarter/Sales Amount", two columns, easy comparison and readability. | • Clarity and comparison<br>• Basic markdown structure. |

The models' ability to arrange sales data and support their formatting choices was investigated with an emphasis on usability. All five models successfully converted the sales data into comprehensible markdown tables. Notably, Gemini's suggestion for an additional "Export to Sheets" capability as it has a smooth integration with the Google ecosystem, adds even more utility to its solution. While Claude worked on finding patterns in the data, DeepSeek worked on creating a refined and well-formatted table. These results indicate how each model found a distinct balance between being easy to use, clear and working well with useful features.

**Results**

Adding to these findings, the method of examining different prompts and how each model responded allowed for a straightforward side-by-side comparison of the models' advantages and disadvantages. The investigation of the five models in this case study led to the following significant results, which are summarized in Figure 4.

| Model | Performance & Accuracy | Ethics & Bias Mitigation | Usability & Integration |
|---|---|---|---|
| DeepSeek | ✓ | ✓ | ✓ |
| ChatGPT | ◐ | ◐ | ✓ |
| Gemini | ✓ | ✓ | ✓ |
| Claude | ✓ | ✓ | ✓ |
| LLaMA | ◐ | ◐ | ✓ |

*Figure 4 Results Analysis of the Case Study Across Five Models*

In situations that dealt with ethical reasoning, DeepSeek consistently produced the most evidence-based answers. Gemini was distinguished by its strong multimodal capabilities and nuanced ethical frameworks and provided answers that were useful in real-world situations. Claude showed good moral reasoning, but its data-driven decision-making was not as deep. Also, although ChatGPT didn't always go into detail on ethical problems, it was a well-rounded program that was easy to use. Lastly, LLaMA put a lot of emphasis on being clear and simple. This is great when you need to be open, but it may not be adequate to meet higher performance or ethical criteria. Furthermore, Table 7 shows a full description of the study results, assessments and suggestions that came from a thorough look of these five LLMs.

Table 7: Evaluation of Gemini, DeepSeek, Claude, GPT and LLaMA on Key Parameters

| Model | Performance and Accuracy | Ethics and Bias Mitigation | Usability and Integration | Insights |
|---|---|---|---|---|
| Gemini | • Strong ability to retrieve factual information.<br>• Structured, evidence-backed as it references IRENA and other sources. | • Adds strong content filtering and ethical reasoning<br>• Includes practical steps for fairness. | • Positive User Feedback<br>• Integration within Google's ecosystem. | • Effective in handling multimodal tasks<br>• High ethical transparency<br>• Strong in real-world examples. |
| DeepSeek | • Superior accuracy in mathematics and technical disciplines.<br>• High context retention, adaptive training and computational efficiency.<br>• Clear, evidence-based summary<br>• Focuses on real-world examples. | • Ethical considerations like bias, content moderation and data transparency<br>• More established mitigation strategies.<br>• Evidence-based with research citations<br>• Shows bias audits. | • High inference speed<br>• Increased calability and energy efficiency. | • Strong in evidence-based reasoning<br>• Data-driven decisions<br>• Professional formatting |
| Claude | • Strong bias management.<br>• High correlation (0.82) with GPT-4 in problem-solving<br>• Concise with good ethical framing<br>• Lacks some evidence | • Built around Anthropic's ethical AI principles<br>• Practical steps<br>• Strong transparency focus<br>• Less evidence | • Balanced between detail and conciseness<br>• Priority over data trends and readability | • Focus on ethical reasoning<br>• Strong bias mitigation<br>• Lacks detailed source citations |
| GPT | • Achieved highest accuracy<br>• Concise, clear<br>• Lacks depth in evidence and examples | • Uses Reinforcement Learning from Human Feedback (RLHF) to better match outputs with human values<br>• High fairness and transparency | Adaptable across various industries | • Balanced in usability<br>• Lacks evidence-backed ethical considerations. |
| LLaMA | • Impressive efficiency<br>• Brief and concise<br>• Lacks in-depth detail and examples. | • Open-source nature<br>• Flexibility and openness<br>• Provides users more control over ethical issues.<br>• Focuses on transparency<br>• Lacks real-world examples. | Focused on clarity and easy comparison | • Good at clarity and simplicity<br>• Lacks in-depth ethical analysis and evidence backing |

In terms of Performance and Accuracy[1], ChatGPT stands out for its ability to understand broad English[7][11]. As it works effectively on a range of tasks, it is a reliable choice for general applications. The analysis showed that Claude is excellent at ensuring factual correctness, despite its tendency to neglect some overall performance in favor of accuracy in particular areas. The ability of Gemini to handle multimodal tasks is one of its unique advantages [19]. It involves text and image data. These are the primary reasons for its high performance and use when these functionalities are required[3]. LLaMA, which is known for its efficiency, performed quite well[4]. This makes it a great option when resource constraints are taken into consideration and optimal performance is needed. Significant improvements in context retention and processing speed were demonstrated by DeepSeek. It demonstrated remarkable accuracy in technical areas in tasks that involve mathematical reasoning and programming [22].

Claude outperforms in terms of ethics and bias mitigation because of its management of bias and good adherence to moral principles. It performs exceptionally well in minimizing hazardous outputs and it is essential for delicate applications. Models like GPT and Gemini also show trustworthy results despite differences in their ability to deal with bias and moral quandaries[4]. LLaMA offers transparency because to its open-source nature[15]. However, it also struggles to work on its biases in the training data. Gemini uses a unique approach that aims to reduce hallucinations while enhancing factual verification[28]. DeepSeek also performs better in this area by using comprehensive mitigating techniques like bias checks.

Each model has unique advantages in terms of use and integration[9][17]. Claude offers a comprehensive, focused strategy for customers that prioritize morals and safety. Gemini offers a seamless experience consumers who are already a part of Google's ecosystem because of its exceptional integration. LLaMA offers a great degree of customisation because of its open-source flexibility which enables developers to alter and adjust the model to match certain needs. Since DeepSeek offers cross-platform compatibility, optimal performance for corporate scaling and powerful inference speed, it is highly useful for demanding technological applications.ng and powerful inference speed, it is highly useful for demanding technological applications.

**Conclusion**

As the field of large language models advances, each iteration improves performance, ethical handling and usability, allowing these models to handle more complicated, real-world circumstances. The differences between the models examined in this study underscore the distinct benefits that each model offers to various activities, pointing out the significance of choosing the appropriate model for a specific use case. Gemini was notable for its multimodal capabilities and sophisticated ethical frameworks, while DeepSeek was exceptional at evidence-based ethical reasoning. Claude showed sound moral reasoning, but his data-driven decision-making was shallow.

Although it occasionally lacked depth in ethical issues, ChatGPT provided a well-rounded performance with great usability. Since LLaMA placed a high value on clarity and simplicity, it works best for open applications but is less useful for intricate ethical or performance requirements.

Looking ahead, future developments in contextual understanding, scalability and the integration of multimodal capabilities will likely open up new avenues for AI to become an even more significant component of our everyday lives and businesses. More human values and AI systems will probably align in the future and better collaboration tools will make sure that these models are useful tools for human decision-making. Ultimately, the next generation of AI models will probably integrate more smoothly into a variety of industries, with a focus on increasing efficiency and scalability.

In conclusion, the variations in performance, ethical practices and usability reflect the unique philosophies behind each model[18]. While some are designed for broad applicability, others focus on specialized tasks or particular use cases. Using each model in situations where it works best and according to the user's demands is the key to realising its full potential.

**Declaration Of Interests**

The authors declare no competing interests.

**Funding Sources**

This research did not receive any specific grant from funding agencies in the public, commercial, or not-for-profit sectors.

**Author contributions**

Urja Kohli: Conceptualization, Data Curation, Formal Analysis, Investigation, Methodology, Resources, Validation, Visualization, Writing – Original Draft. Aditi Singh: Data Curation, Validation, Visualization, Writing – Original Draft. Arun Sharma: Formal Analysis, Project Administration, Supervision, Writing – Review & Editing

# References

[1] Khlaif, Z. N., Mousa, A., Hattab, M. K., Itmazi, J., Hassan, A. A., Sanmugam, M. and Ayyoub, A. (2023). The potential and concerns of using AI in scientific research: ChatGPT performance evaluation. JMIR Medical Education, 9, e47049.

[2] Hua, S., Jin, S. and Jiang, S. (2024). The limitations and ethical considerations of chatgpt. Data intelligence, 6(1), 201-239.

[3] Rane, N., Choudhary, S. and Rane, J. (2024). Gemini versus ChatGPT: applications, performance, architecture, capabilities and implementation. Journal of Applied Artificial Intelligence, 5(1), 69-93.

[4] Touvron, H., Lavril, T., Izacard, G., Martinet, X., Lachaux, M. A., Lacroix, T., ... and Lample, G. (2023). Llama: Open and efficient foundation language models. arXiv preprint arXiv:2302.13971.

[5] Lin, S., Hilton, J. and Evans, O. (2021). Truthfulqa: Measuring how models mimic human falsehoods. arXiv preprint arXiv:2109.07958.

[6] Zuccon, G. and Koopman, B. (2023). Dr ChatGPT, tell me what I want to hear: How prompt knowledge impacts health answer correctness. arXiv preprint arXiv:2302.13793.

[7] Kalyankar, H. B., Taubert, L. and Wygnanski, I. J. (2018). Re-orienting the Turbulent flow over an Inclined Cylinder of Finite Aspect ratio. In 2018 Flow Control Conference (p. 4026).

[8] Lim, J., Leinonen, T., Lipponen, L., Lee, H., DeVita, J. and Murray, D. (2023). Artificial intelligence as relational artifacts in creative learning. Digital Creativity, 34(3), 192-210.

[9] Andersson, M. and Marshall Olsson, T. (2023). ChatGPT as a Supporting tool for System Developers: Understanding User Adoption.

[10] Schneider, P., Klettner, M., Simperl, E. and Matthes, F. (2024). A Comparative Analysis of Conversational Large Language Models in Knowledge-Based Text Generation. arXiv preprint arXiv:2402.01495.

[11] Antaki, F., Touma, S., Milad, D., El-Khoury, J. and Duval, R. (2023). Evaluating the performance of ChatGPT in ophthalmology: an analysis of its successes and shortcomings. Ophthalmology science, 3(4), 100324.

[12] Gao, C.A., Howard, F.M., Markov, N.S. et al. Comparing scientific abstracts generated by ChatGPT to real abstracts with detectors and blinded human reviewers. npj Digit. Med. 6, 75 (2023). https://doi.org/10.1038/s41746-023-00819-6

[13] Imran, M. and Almusharraf, N. (2024). Google Gemini as a next generation AI educational tool: a review of emerging educational technology. Smart Learning Environments, 11(1), 22.

[14] Saeidnia, H. R. (2023). Welcome to the Gemini era: Google DeepMind and the information industry. Library Hi Tech News, (ahead-of-print).

[15] Jiao, J., Afroogh, S., Xu, Y. and Phillips, C. (2024). Navigating llm ethics: Advancements, challenges and future directions. arXiv preprint arXiv:2406.18841.

[16] Borji, A. and Mohammadian, M. (2023). Battle of the wordsmiths: Comparing chatgpt, gpt-4, claude and bard. GPT-4, Claude and Bard (June 12, 2023).

[17] Pokorný, J. (2023, November). Data Integration in a Multi-model Environment. In *International Conference on Information Integration and Web Intelligence* (pp. 121-127). Cham: Springer Nature Switzerland.

[18] Chitimoju, S. (2024). The Evolution of Large Language Models: Trends, Challenges and Future Directions. Journal of Big Data and Smart Systems, 5(1).

[19] Team, G., Anil, R., Borgeaud, S., Alayrac, J. B., Yu, J., Soricut, R., ... and Blanco, L. (2023). Gemini: a family of highly capable multimodal models. arXiv preprint arXiv:2312.11805.

[20] Vavekanand, R. and Sam, K. (2024). Llama 3.1: An in-depth analysis of the next-generation large language model.

[21] Deng, J. and Lin, Y. (2023). The benefits and challenges of ChatGPT: An overview. Benefits, 2(2), 2022.

[22] Chowdhury, M. N. U. R., Haque, A. and Ahmed, I. DeepSeek vs. ChatGPT: A Comparative Analysis of Performance, Efficiency and Ethical AI Considerations.

[23] Cheng, L., Hu, M. and Hong, T. (2025). Profiling Elements, Risks and Governance of Artificial Intelligence: Implications from DeepSeek. *International Journal of Digital Law and Governance*, (0).

[24] Boomuang, J. and Wang, F. Balancing Performance and Ethics: Responsible AI Strategies for DeepSeek-R1-Powered Agent.

[25] Joshi, S. (2025). A Comprehensive Review of DeepSeek: Performance, Architecture and Capabilities.

[26] Singh, S., Bansal, S., Saddik, A. E. and Saini, M. (2025). From ChatGPT to DeepSeek AI: A Comprehensive Analysis of Evolution, Deviation and Future Implications in AI-Language Models. *arXiv preprint arXiv:2504.03219*.

[27] Li, Z., Fan, Z., Chen, J., Zhang, Q., Huang, X. J. and Wei, Z. (2023, July). Unifying cross-lingual and cross-modal modeling towards weakly supervised multilingual vision-language pre-training. In Proceedings of the 61st Annual Meeting of the Association for Computational Linguistics (Volume 1: Long Papers) (pp. 5939-5958).

[28] Vaillancourt, E. and Thompson, C. (2024). Instruction tuning on large language models to improve reasoning performance. Authorea Preprints.

[29] Rathod, V., Nabavirazavi, S., Zad, S. and Iyengar, S. S. (2025, January). Privacy and security challenges in large language models. In 2025 IEEE 15th Annual Computing and Communication Workshop and Conference (CCWC) (pp. 00746-00752). IEEE.

[30] From open-ended to multiple-choice: evaluating diagnostic performance and consistency of ChatGPT, Google Gemini and Claude AI / Y. O. Mykhalko, Y. F. Filak, Y. V. Dutkevych-Ivanska, M. V. Sabadosh, Y. I. Rubtsova // Wiadomości Lekarskie Medical Advances. – 2024, – Vol. 77(10). – p. 1852-1856.

[31] Hu, Y., Song, K., Cho, S., Wang, X., Foroosh, H., Yu, D. and Liu, F. (2024). Can Large Language Models do Analytical Reasoning?. *arXiv preprint arXiv:2403.04031*.

[32] Neuman, W. R., Coleman, C. and Shah, M. (2025). Analyzing the Ethical Logic of Six Large Language Models. *arXiv preprint arXiv:2501.08951*.